\documentclass[letterpaper]{article} 
\usepackage{aaai24}  
\usepackage{times}  
\usepackage{helvet}  
\usepackage{courier}  
\usepackage[hyphens]{url}  
\usepackage{graphicx} 
\usepackage{natbib}  
\usepackage{caption} 
\usepackage{xcolor}
\usepackage{algorithm}
\usepackage{algorithmic}

\usepackage{newfloat}
\usepackage{listings}
\DeclareCaptionStyle{ruled}{labelfont=normalfont,labelsep=colon,strut=off} 
\floatstyle{ruled}
\newfloat{listing}{tb}{lst}{}
\floatname{listing}{Listing}
\nocopyright

\usepackage{amsfonts}
\usepackage{amsmath}
\usepackage{booktabs}
\title{Image Safeguarding: Reasoning with Conditional Vision Language Model \\and Obfuscating Unsafe Content Counterfactually 
}

\author {
    Mazal Bethany\equalcontrib\textsuperscript{\rm 1 \rm 2},
    Brandon Wherry\equalcontrib\textsuperscript{\rm 1 \rm 2},
    Nishant Vishwamitra\textsuperscript{\rm 1},
    Peyman Najafirad\textsuperscript{\rm 1 \rm 2} \thanks{Corresponding author}
}
\affiliations {
    \textsuperscript{\rm 1}University of Texas at San Antonio
    \textsuperscript{\rm 2}Secure AI and Autonomy Lab\\
    \{mazal.bethany, brandon.wherry, nishant.vishwamitra, peyman.najafirad\}@utsa.edu
}

\begin{document}

\maketitle

\begin{abstract}
Social media platforms are being increasingly used by malicious actors to share unsafe content, such as images depicting sexual activity, cyberbullying, and self-harm. Consequently, major platforms use artificial intelligence (AI) and human moderation to obfuscate such images to make them safer. Two critical needs for obfuscating unsafe images is that an accurate rationale for obfuscating image regions must be provided, and the sensitive regions should be obfuscated (\textit{e.g.} blurring) for users' safety. This process involves addressing two key problems: (1) the reason for obfuscating unsafe images demands the platform to provide an accurate rationale that must be grounded in unsafe image-specific attributes, and (2) the unsafe regions in the image must be minimally obfuscated while still depicting the safe regions. In this work, we address these key issues by first performing visual reasoning by designing a visual reasoning model (VLM) conditioned on pre-trained unsafe image classifiers to provide an accurate rationale grounded in unsafe image attributes, and then proposing a counterfactual explanation algorithm that minimally identifies and obfuscates unsafe regions for safe viewing, by first utilizing an unsafe image classifier attribution matrix to guide segmentation for a more optimal subregion segmentation followed by an informed greedy search to determine the minimum number of subregions required to modify the classifier's output based on attribution score. Extensive experiments on uncurated data from social networks emphasize the efficacy of our proposed method. We make our code available at: https://github.com/SecureAIAutonomyLab/ConditionalVLM

\end{abstract}

\section{Introduction}
\label{introduction}
Social media is being increasingly misused by bad actors to share sexually explicit, cyberbullying, and self-harm content~\cite{cyberbullying_increase_12news,chelmis2019minority,adler2022tornado}. However, social media platforms are required by law to safeguard their users against such images~\cite{exon1996communications}, as well as provide a rationale for why such images are flagged~\cite{gesley2021germany,cabral2021eu} for the purpose of transparency. In response, major platforms have deployed AI and human-based content moderation techniques to flag and obfuscate (\textit{i.e}, make the image safer by blurring sensitive regions) such images~\cite{bethany2023towards}. This process involves obfuscating (\textit{e.g.} by blurring or blocking) unsafe image regions in the image~\cite{li2017effectiveness} along with generating a rationale that backs up the decision to obfuscate the flagged images~\cite{appealedcontent}. \looseness=-1

The image obfuscation process faces two critical problems regarding how much of the unsafe image is obfuscated and why it is obfuscated: 
\textit{First}, the decision to deem an image unsafe and obfuscate it demands providing a rationale for the decision. For example, Instagram moderators are required to provide a legal rationale~\cite{bronstein2021deplatforming,are2020instagram} to back up their decision~\cite{instagramrationalemoderator}. 
Existing visual reasoning methods~\cite{li2022mplug,li2023blip,dai2023instructblip} are severely limited for unsafe images such as sexually explicit, cyberbullying, and self-harm since they cannot provide a rationale grounded in attributes that are specific to such images, such as rude hand gestures in cyberbullying images~\cite{vishwamitra2021towards}, or sensitive body parts in sexually-explicit images~\cite{facebook_tool_revenge_porn}. 
\textit{Second}, the unsafe image needs minimal obfuscation while still depicting the safe regions for evidence collection and investigation~\cite{facebookalwayshuman}. For instance, human moderators need to determine the age of the person in the image (\textit{e.g.}, in child sexual abuse material (CSAM) investigations), look for identifiers (\textit{e.g.}, tattoos,  scars, and unique birthmarks), and determine their location information (\textit{e.g.}, landmarks, geographical features, and recognizable surroundings).
Current segmentation techniques~\cite{chandrasekaran2021combinatorial,vermeire2022explainable,bethany2023towards} cannot minimally identify the regions and consequently impede investigations that pertinently need full details of the remaining safe regions.
\looseness=-1


In this work, we take the first step towards addressing a pertinent, but overlooked problem of the \emph{image moderation process} in social media platforms. 
%
Our major objective is to first identify and minimally obfuscate the sensitive regions in an unsafe image such that the safe regions are unaltered to aid an investigation, and then provide an accurate rationale for doing so, that is grounded in unsafe image attributes (\textit{e.g.}, private body parts, rude gestures or hateful symbols). 
To this end, we address this problem in two steps: (1) we develop a novel unsafe image rationale generation method called ConditionalVLM (\textit{i.e.}, conditional vision language model) that leverages the state-of-the-art large language models (LLM)-based vision language models~\cite{fang2023eva} to perform an in-depth conditional inspection to generate an accurate rationale that is grounded in unsafe image attributes; and (2) minimally obfuscating the sensitive regions \textit{only} by calculating the classifier attribution matrix using a FullGrad-based model~\cite{srinivas2019full} and then utilize this information to guide Bayesian superpixel segmentation~\cite{uziel2019bayesian} for a more informed and optimal dynamic subregion segmentation, via calculating the attribution score of each subregion. Finally, we utilize a discrete optimization technique such as informed greedy search to determine the minimum number of subregions required to modify the classifier's output, using the score attribution.

Our work has profound implications for the safety of social media content moderators, by greatly reducing their need to view unsafe content~\cite{steiger2021psychological}, social media users who are minors or sensitive to such content~\cite{hargrave2009harm}, and law enforcement agents who need to investigate such images as part of their investigation~\cite{krause2009identifying}. We make the following contributions:

\begin{itemize}
\item We develop ConditionalVLM, a visual reasoning model that generates accurate rationales for unsafe images by leveraging state-of-the-art VLMs conditioned on pre-trained unsafe image classifiers.

\item We develop a novel unsafe image content obfuscation algorithm that minimally obfuscates only the unsafe regions while keeping the rest of the image unaltered for investigations.

\item Evaluations of our work show that it can categorize the three social media unsafe categories of images with an accuracy of 93.9\%, and minimally segment only the unsafe regions with an accuracy of 81.8\%.

\end{itemize}





\section{Related Works}
\label{related_works}


\begin{figure*}[t!]
    \centering
    \includegraphics[width=11cm]{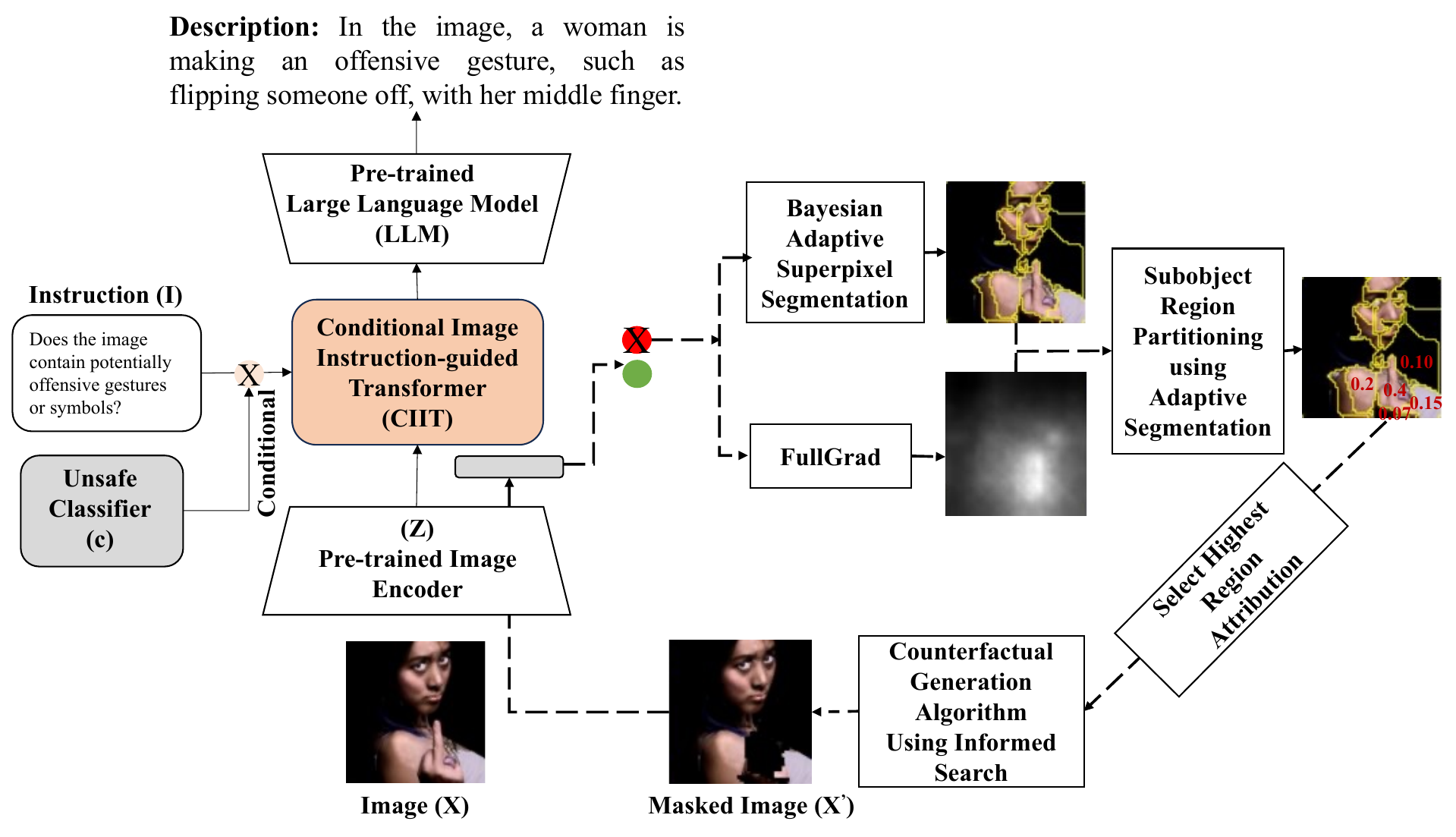}
    \caption{Overview of the proposed architecture. The initial module utilizes ConditionalVLM for classifying images as safe or unsafe, while the subsequent module proposes counterfactual visual explanations to identify and obfuscate the unsafe regions within the image.}
    \label{fig:overview}
\end{figure*}

\subsection{Safeguarding Images} 
Social media platforms are constantly misused to share unsafe content. For example, sexually-explicit images~\cite{ashurst2015young,sanchez2019practitioner}, non-consensual intimate images (NCII)~\cite{lenhart2016nonconsensual} and child sexual abuse material (CSAM) are being increasingly shared on major social media platforms~\cite{sanchez2019practitioner}. Cyberbullying, a critical issue affecting adolescents and adults is spread unabated via images~\cite{vishwamitra2021towards}. Furthermore, self-harm images are widely spread on these platforms that further alienate vulnerable users~\cite{john2018self}. To defend against this threat, social media platforms alter the unsafe image by blurring the sensitive regions. This obfuscation of such unsafe images has substantial implications. For instance, social media platforms employ over a million moderators globally~\cite{bbc2021facebook}, who manually view unsafe images which is known to have an adverse effect on their mental health, including PTSD~\cite{reuters2021judge}. Vulnerable users, such as minors also need image safeguarding methods that can shield them from unsafe content. Furthermore, law enforcement agents who investigate images from a crime scene need image safeguarding since such images contain extremely disturbing images, such as CSAM. These applications have a critical need, the unsafe regions must be minimally obfuscated. For instance, a moderator needs to view the safe regions to conduct investigations, such as determining the age of the person to report CSAM content to law enforcement. 




\subsection{Vision-Language Models}

Given the achievements of pre-trained models in computer vision (CV) and natural language processing (NLP), numerous studies have attempted to pre-train large-scale models that incorporate both vision and language modalities. These models are commonly referred to as Vision-Language Models (VLM). One group of methods propose end-to-end approaches for learning vision-language models. Works such as CLIP~\cite{radford2021learning} use a contrastive pre-training process to jointly train an image and a text encoder on image-text pairs. Other works such as BEIT-3 \cite{wang2023image} use multiway transformers for general-purpose modeling and carry out masked modeling on images, text, and image-text pairs. Some methods use modular techniques to utilize pre-existing models to interpret image data and take advantage of established LLMs. 
One major hurdle in these works is coordinating visual features within the realm of text. To accomplish this, works such as Flamingo \cite{alayrac2022flamingo} add cross attention layers and train these layers on image-text pairs. Works such as BLIP-2 use a frozen image encoder and LLM and propose a querying transformer to bridge modalities \cite{li2023blip}. Other methods such as LENS \cite{berrios2023towards} require no additional training, and develop visual vocabularies of an image by collecting tags, attributes and captions from various models. LENS then uses this visual vocabulary to generate text prompts by which questions can be asked about an image on a frozen LLM. A major limitation of these  methods is that they do not have conditioning capability~\cite{ramesh2022hierarchical}, a crucial requirement to ground the output in domain-specific attributes.\looseness=-1

\subsection{Image Segmentation and Counterfactual Explanation for Obfuscation}

Another type of explanation that is growing in popularity due to its ability to address several of these issues is counterfactual explanations~\cite{wachter2017counterfactual}. 
A counterfactual explanation can be defined as taking the form: a decision $y$ was produced because variable $X$ had values ($v1, v2, \dots$) associated with it. If $X$ instead had values ($v1', v2', \dots$), and all other variables had remained constant, score $y'$ would have been produced. 
Some works such as BEN \cite{chandrasekaran2021combinatorial}, SEDC \cite{vermeire2022explainable}, and CSRA \cite{bethany2023towards} have explored region-based counterfactual visual explanations. However, existing approaches face two key challenges: 1. suboptimal subregion boundaries, leading to excessive parts of the image being identified as causing a decision, and 2. high time complexity \(2^K\) in searching for a counterfactual in an image of K regions. BEN and SEDC segment an input image into K static subregions without any prior knowledge of the classifier, resulting in an uninformed search strategy for finding the counterfactual examples. While CSRA does use prior knowledge of the classifier to inform the search of the counterfactual example, BEN, SEDC and CSRA do not jointly optimize the subregions boundaries and minimize the number of subregions, which is particularly important for obfuscation applications where preserving as much context as possible is preferred.

\section{Method}

Figure \ref{fig:overview} illustrates the architecture of our proposed approach, which consists of two modules. The initial module proposes a conditional visual language model designed for image reasoning. The model classifies images as safe or unsafe by understanding the interactions or activities of entities within the image, using its comprehension of visual features and linguistic annotations. In the subsequent module, counterfactual visual explanations are proposed to precisely identify sub-object regions of the image contributing to its unsafe classification for obfuscation.

\subsection{Conditional Vision-Language Model}

We introduce a framework that synergistically combines the strengths of large language models (LLMs) with the specific requirements of large image encoders. Additionally, it provides more explicit control over visual features being reasoned. The ConditianalVLM architecture is anchored by three pivotal components, as depicted in Figure \ref{fig:overview}:

\noindent \textbf{A Large Pre-trained Image Encoder} takes an image X as input and outputs a visual embedding representation of the image, \(Z = g(X)\). We explore a state-of-the-art pre-trained vision transformer ViT-g/14 from EVA \cite{fang2023eva}. 

\noindent \textbf{A Conditional Image Instruction-guided Transformer (CIIT)} employs contrastive language-image pre-training to encode visual data in congruence with a specific language prompt. Additionally, we condition this language prompt using pre-trained unsafe image classifiers. This allows the model to match and parse the unsafe visual embedding effectively, while also providing more explicit control over unsafe visual features \cite{ramesh2022hierarchical}. 
CIIT utilizes a pre-trained Q-Former model \cite{li2023blip}, which is conditioned on image classifiers as control code \(c\) on unsafe image content such as sexually explicit, cyberbullying, and self-harm. \looseness=-1
\begin{itemize}
    \item A prior \(p(I|c)\) that produces CIIT instruct prompt \(I\) conditioned on control code \(c\).
    \item A transformer decoder \(p(L|I
, c)\) that produces contrastive embedding \(L\) conditioned on Instruct prompt \(I\) and control code \(c\).
\end{itemize}

The transformer decoder allows us to invert images given their CIIT Instruct prompt, while the prior allows us to learn
a generative model of the image embeddings themselves. Taking the product of these two components yields a generative
model \(P(L|c)\) of embedding \(L\) given control \(c\):

\begin{equation}
p(L|c) = p(L, I|c) = p(L|I,c)p(I|c) 
\end{equation}

The control code \(c\) provides a point of control over the CIIT generation process. The distribution can be decomposed using the chain rule of probability and trained with a loss that takes the control code into account.

\begin{equation}
    p(L|c) = \prod^{n}_{i=1} p(L_i|L_{<i},c)
\end{equation}

We train the model with parameters \(\theta\) to minimize the negative log-likelihood over a dataset \(D = {X_1, ..., X_n}\):

\begin{equation}
    \mathcal{L}(D)=-\sum^{|D|}_{k=1} \log p_{\theta}(L^k_i|L_{<i},c^k)
\end{equation}

\noindent \textbf{A Pre-trained Large Language Model Decoder}
takes a text embedding \(L\) as input and outputs linguistic sentences derived from the embedding, \(Text = LLM(L)\). We choose Vicuna \cite{Vicuna2023} as our LLM decoder which is constructed upon LLaMA \cite{touvron2023llama} and can perform a wide range of complex linguistic tasks.





\subsection{Counterfactual Subobject Explanations for Obfuscation}
In order to connect region attribution to provide counterfactual subobject region explanation of an image, relative to a given machine learning predictive model, we propose a two-phase approach. The pipeline of the proposed approach is illustrated in {Figure~\ref{fig:overview}. We first partition the image into non-intersecting subobject regions and measuring region attribution value to each region using gradient attribution maps in Section 3.1 and Section 3.2. The counterfactual analysis of alternate versions of the image using a greedy search algorithm using regions with highest attribution values for counterfactual analysis is followed in Section 3.3. 


\noindent \textbf{Subobject Region Partitioning using Adaptive Segmentation.}
We represent a given image,\(X\)as a non-intersecting set of \(K\) regions given by \(\{z_1, z_2, \cdots , z_K\}\). The boundaries of these regions are defined by clustering algorithms that use color and spatial information and are called superpixels. Let \(Z\) represent the \(K\) region segmentation, \(z_i\) represent the label assigned to \(X_i\) and \(j\) represent the label of some arbitrary cluster. An image must be segmented into meaningful subobject regions in order to allow for a counterfactual analysis of the image by the binary predictive model \(f (X) \rightarrow {0, 1}\). These regions serve as the features that are analyzed in the counterfactual analysis. To maximize the efficiency of a counterfactual analysis, we require an adaptive segmentation method. Many segmentation methods are wasteful in their assignment of many segments to uninformative regions, while not segmenting detailed regions enough. Such a method should be able to respect pixel connectivity and spatial coherence and requires an adaptive number of regions. K-means based clustering methods are a fast and simple basis for leading segmentation, however Gaussian Mixture Models (GMM) may be better suited for an adaptive segmentation method since we need to capture the heterogeneity in the pixel distribution of various types of images.\looseness=-1

Let \(N=h*w\) be the number of pixels in an image, \(X\) with \(c\) color channels. The values attributed to the pixels in  \(X\) can be denoted as \(X_i = (l_i, c_i) \in \mathbb{R}^ 5\), where \(l_i \in \mathbb{R}^2\) represent the \(x,y\) coordinate location and \(c_i \in \mathbb{R}^3\) represent the RGB color information. Superpixel clustering methods with spatial coherence aim to partition \((X_i)^N_{i=1}\) into \(K\) disjoint groups. Let \(Z\) represent the \(K\) region segmentation, \(z_i\) represent the label assigned to \(X_i\) and \(j\) represent the label of some arbitrary cluster. Where  \(\mathcal{N}(X;\mu_j, \Sigma_j)\) is a Gaussian PDF with mean \(\mu_j\) and a covariance matrix \(\Sigma_j\) of size \(n*n\), the PDF of a GMM with \(K\) components is

\begin{equation*} 
    p(X;(\mu_j, \Sigma_j, \lambda_j)^K_{j=1}) = \sum_{j=1}^{K} \lambda_j \mathcal{N}(X|\mu_j, \Sigma_j)
\end{equation*}

The mixing coefficients \(\lambda_j\) in the PDF of a GMM form a convex combination where: \begin{equation*} \sum_{j=1}^{K} \lambda_j =1, \lambda_j \ge 0 \quad \forall j \end{equation*}
and this allows for a globally optimal clustering. Given a Gaussian distribution \(j\) where \(\theta_j = (\mu_j , \Sigma_j )\), a Bayesian GMM with random variables \((\theta_j)^K_{j=1}\) and \((\lambda_j)^K_{j=1}\) are drawn from \(p((\theta_j, \lambda_j)^K_{j=1})\), a prior distribution. Assuming independence, the prior distribution can be factorized as follows

\begin{equation*} \label{eq:gmm1}
    p((\theta_j, \lambda_j)^K_{j=1}) = p((\lambda_j)^K_{j=1}) \prod_{j=1}^{K} p(\theta_j)
\end{equation*}

Using a Normal-Inverse Wishart (NIW) for \(p(\theta_j)\) and a Dirichlet distribution for \(p((\lambda_j)^K_{j=1})\) gives us posterior distributions in the same form as the priors. Furthermore, the updates from the priors are given
in closed form.

The Bayesian GMM inference to calculate \(Z\) can be done by performing Gibbs sampling, alternating between the following equations:

\begin{equation*} \label{eq:gmm2}
    p((\lambda_j)^K_{j=1}|Z, (X_i)^N_{i=1}) \prod_{j=1}^{K} p((\theta_j, \lambda_j)|Z, (X_i)^N_{i=1})
\end{equation*}

\begin{equation*} \label{eq:gmm3}
    p(Z|(\theta_j, \lambda_j)^K_{j=1}, (X_i)^N_{i=1})
\end{equation*}

\noindent \textbf{Subobject Region Attribution Value.} \label{ssec:avgatt} We start by creating the FullGrad \cite{srinivas2019full} attribution map for image feature attribution. Given an image \(X\) and the feature maps generated by the FullGrad \(L[u,v]\) of width u and height v for the model prediction, the goal of the visual attention model is to identify the discriminative regions of the image that significantly influence the class prediction score of the predictive model using \(L[u,v]\) pixel attribution values.


The attribution map of the FullGrad method is generated by propagating an image through a CNN, obtaining the output score before the softmax layer, and then computing the gradients with respect to the input (input-gradients) and the biases at each layer (bias-gradients). These gradients are then combined, with each bias-gradient reshaped to match the input dimensionality and all gradients summed to form the FullGrad attribution map.

FullGrad Definition: \emph{Consider a CNN model f, with x denoting the input and b denoting the biases at each layer, c representing the channels of layer k. Furthermore, given an output of interest f(x), and a postprocessing operator \(\psi(\cdot)\) the FullGrad attribution map \(L_{FullGrad}\) is defined as:}

\begin{equation*} \label{eq:gradcam1}
    L_{FullGrad} = \psi (\nabla_x f(x) \odot x) + \sum_{k \in K} \sum_{c \in c_k} \psi (f^b(x)_c) 
\end{equation*} 




To facilitate an efficient sampling of regions in the counterfactual analysis, we utilize the FullGrad attribution map.

\medskip 

\textbf{Definition 1: (Subobject Region Attribution Score)} \textit{Using the attribution map of model \(f(X)\) and the subobject regions \(\{z_1, z_2, \cdots , z_K\}\) created by adaptive segmentation for the input image X, we define the subobject region attribution score,  \(\{s_1, s_2, \cdots , s_K\}\) as follows: 
}
\begin{equation*} \label{eq:regionscore}
   s_k = \frac{1}{n.m} \sum_{n}\sum_{m} L_{FullGrad(F,X)} [i,j], X[i,j]\in z_k\
\end{equation*}

Although feature attributions highlight features that are
significant in terms of how they affect the model’s ability to predict, they do not indicate that altering significant features would result in a different desired outcome. 

\medskip 

\textbf{Definition 2: (Subobject Region Confidence Reduction) } \textit{Given a model \(Y=f(X)\) that takes an image X with subobject regions \(X=[z_0,z_1,...,z_n]^T\) and outputs a probability distribution Y. The confidence reduction \(cr_k\) of subobject region \(z_k\), \((k \in [1,n])\) towards Y is the change of the output by masking the k-th subobject region of \(X\), while being classified as the same 
class, as follows:}

\begin{equation*} \label{eq:confidence}
   cr_k = f(X) - f(X \circ Mask(z_k))
\end{equation*}

In Sec 3.3, we present our greedy region search algorithm which utilizes subobject region attribution score as heuristics and employs confidence level for causal obfuscation using counterfactual subobject region explanations.



\noindent \textbf{Counterfactual Generation Using Informed Subobject Region Search.}
The previous sections lead us to the minimum region masking problem.
This can be computationally expensive to solve, as it requires the masking and analysis of \(2^K\) different combination of regions, \(Z\) of \(X\) based on Section 3.1 . Rather than solving the problem directly, we find an approximate solution using a greedy region search. 

Given a predictive model \(f : X \rightarrow {0, 1}\), we can define the set of counterfactual explanations for an input \(x \in X\) as \(x’\) while \(arg min \, x’ d (x, x’)\) and \(x’= \{x \in X | f(x) \neq f(x')\}\). In other words \(x’= \{x \in X | f(x) \neq f(x')\}\)  , contains all the inputs x for which the model f returns a prediction different from \(f(x)\) while minimizing the distance between \(x\) and \(x’\).\looseness=-1



Our greedy region search, starts with us first sorting the \(K\) regions in descending order by the average attribution for each region which were calculated in subsection \ref{ssec:avgatt}. The greedy region search considers a subset of regions \(k\in K\). \(k\) begins with the top 
region by average attribution and iteratively expands to the top two regions by average attribution and so on until an \(x'\) is found such that \(f(x') \neq f(x)\).



\section{Experimental Evaluation}
\label{experiments}


\subsection{Datasets}

We evaluated our Conditional VLM and Counterfactual Subobject Explanation methods on three datasets of real-world harmful images to study the practical application of counterfactual subobject explanations. 

\noindent \textbf{Sexually Explicit:} First, we sampled a subset of images from an NSFW images dataset~\cite{alagiri2021nsfw} consisting of 334,327 images by selecting the ``porn'', ``neutral'', and ``sexy'' classes. We combine the ``neutral'', and ``sexy'' classes into a single class of ``safe'' images. 

\noindent \textbf{Cyberbullying:}
Second, we used a cyberbullying images~\cite{vishwamitra2021towards} dataset consisting of nearly 20,000 images belonging to the classes ``cyberbullying'' and ``non-cyberbullying''. 

\noindent \textbf{Self-Harm:}
Third, we used a self-harm images dataset \cite{bethany2023towards}, consisting of 5000 images with classes ``self-harm'' and ``non self-harm''.

\begin{figure}[b!]
    \centering
    \includegraphics[width=5cm]{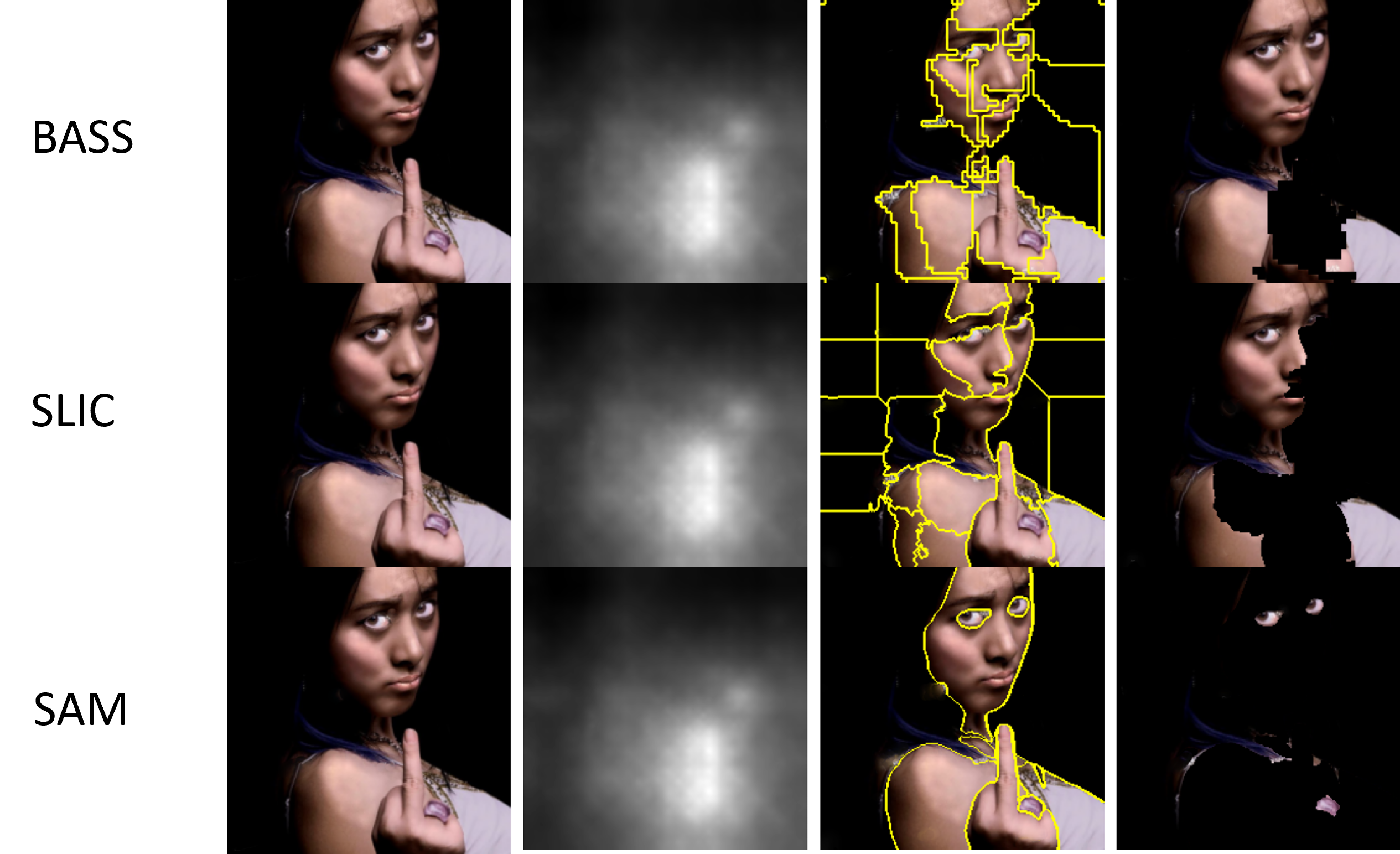}
    \caption{Examples of segmentation methods on a cyberbullying image. From top to bottom: (1) BASS, (2) SLIC, (3) SAM.}
    \label{fig:partitions}
\end{figure}

\subsection{Evaluation Settings}


\noindent \textbf{ConditionalVLM.}
We compare our proposed method against other state-of-the-art image-to-text models such as InstructBLIP \cite{dai2023instructblip}, OFA-Large \cite{wang2022ofa}, and mPLUG \cite{li2022mplug}. We use the implementations of these methods from HuggingFace. For InstructBLIP, we use InstructBLIP-Vicuna-13b with num\_beams=5, max\_length=512, min\_length=1, top\_p=0.9, repetition\_penalty=1.5, length\_penalty=1.0, and temperature=1. The image encoder for this implementation of InstructBLIP was Vit-g/14 \cite{fang2023eva}. For mPLUG, we use the parameters do\_sample=True, top\_k=5, and max\_length=512. For OFA we use the parameters of num\_beams=5, no\_repeat\_ngram\_size=3. To demonstrate our ConditionalVLM framework, we modify the InstructBLIP-Vicuna-13b architecture to include a CIIT, which we call ConditionalBLIP. All experiments were carried out on a DGX 8x A100 GPU, with 80GB of VRAM each.\looseness=-1

We fine-tuned a ResNet-50 classifier available in Pytorch \cite{NEURIPS2019_9015} using pre-trained model weights trained from the ImageNet dataset \cite{deng2009imagenet}. The NSFW, cyberbullying and self-harm datasets were each divided into train, validation, and test sets, with 80\% being allocated to the train set, and 10\% each allocated to validation and test sets. We trained the models for 50 epochs and selected the models that have the highest classification accuracies on the validation sets. These models achieved accuracies of 98.9\%, 91.9\% and 97.6\% respectively on the test set in our experiments. We use these classifiers as the control code for the CIIT in ConditionalBLIP.

\noindent \textbf{Counterfactual Subobject Explanations for Obfuscation.}
To test different segmentation methods, we experimented with SLIC \cite{achanta2010slic}, Felzenszwalb \cite{felzenszwalb2004efficient}, and Compact Watershed \cite{neubert2014compact} segmentation methods implemented in the scikit-image library~\cite{scikit-image}, Segment Anything Model (SAM) \cite{kirillov2023segment}, and Bayesian Adaptive Superpixel Segmentation \cite{uziel2019bayesian}. For our experiments, we selected the following parameters for each segmentation method: for SLIC, we chose the number of segments to be 25 and compactness equal to 1; for Felzenszwalb we selected the scale to be 500, sigma to be 0.5, and a minimum component size of 200; for Compact Watershed, we chose the number of markers to be 25 and the compactness parameter to be 0.001.

We used the following attribution map methods in our experiments: (Grad-CAM \cite{selvaraju2017grad}, XGrad-CAM \cite{fu2020axiom}, Grad-CAM ++ \cite{chattopadhay2018grad}, FullGrad \cite{srinivas2019full}, and Ablation-CAM \cite{ramaswamy2020ablation}). For the implementation of the attribution map methods, we use the Pytorch Grad-CAM library~\cite{jacobgilpytorchcam}.

\subsection{Evaluation Metrics}

\noindent \textbf{ConditionalVLM.}
We evaluate VLM's ability to investigate three different unsafe image categories in two phases. In the first phase, we conduct a coarse-grained evaluation by having human evaluators determine based off of the image descriptions produced by the VLM whether a moderator should be able to understand which dataset of unsafe image the image belongs to. In this evaluation, a team of three human evaluators who were involved in this research were asked to evaluate whether these descriptions produced by the VLM on the questions of ''What is happening in the image?'', and ''What are the people doing?'' were sufficient to accurately categorize them into the correct dataset that the unsafe image the image belongs to. The final labels were assigned by majority voting.

In the second phase, we conduct a fine-grained evaluation by having human evaluators evaluate the responses of the VLM to curated moderator questions with respect to an unsafe image image. These fine-grained questions ask about specific attributes of images relating to the unsafe image categories. In this evaluation, the same team of evaluators were asked to determine whether the answers produced by the VLM correctly answered these curated questions.

\begin{table}[b]
\centering
\resizebox{0.4\textwidth}{!}{
\begin{tabular}{llp{4cm}}
\toprule
\textbf{Dataset}           & \textbf{Model}           & \textbf{Human Evaluation} \\ \midrule
                  & \textbf{ConditionalBLIP} & \textbf{94.5}                                                                         \\ 
Sexually Explicit & InstructBLIP    & 80.0                                                                         \\ 
                  & mPLUG           & 73.5                                                                         \\ 
                  & OFA-Large       & 13.5                                                                         \\ \midrule
                  & \textbf{ConditionalBLIP} & \textbf{99.5}                                                                         \\ 
Cyberbullying     & InstructBLIP    & 23.0                                                                         \\ 
                  & mPLUG           & 16.0                                                                         \\ 
                  & OFA-Large       & 10.0                                                                         \\ \midrule
                  & \textbf{ConditionalBLIP} & \textbf{99.0}                                                                        \\ 
Self-Harm         & InstructBLIP    & 47.0                                                                         \\ 
                  & mPLUG           & 51.5                                                                         \\ 
                  & OFA-Large       & 17.0                                                                         \\ \bottomrule
\end{tabular}
}
\caption{Quantitative results of VLM on coarse-grained image descriptions.}
\label{tab:pass1}
\end{table}

\begin{table}[t]
\centering
\resizebox{0.9\columnwidth}{!}{%
\begin{tabular}{lp{4cm}p{2cm}p{2.5cm}}
\toprule
\textbf{Dataset}           & \textbf{Questions}                                                                      & \textbf{InstructBLIP} & \textbf{ConditionalBLIP} \\ \midrule
                  & Does the image contain any sexual content or nudity?                           & 61.1                                                & 99.2                                          \\ \cline{2-4}
Sexually Explicit & Does the image contain explicit sexual content?                                & 37.9                                                & 99.9                                         \\ \midrule
                  & Does the image contain potentially offensive gestures or symbols?              & 59.1                                                & 80.0                                          \\ \cline{2-4}
                  & Does the image contain any weapons?                                            & 99.2                                                & 94.0                                          \\ \cline{2-4}
Cyberbullying     & Is there any aggressive behavior shown in the image?                           & 56.4                                                & 98.3                                          \\ \cline{2-4}
                  & Is there any text in the image that might be considered hate speech?           & 57.2                                                & 95.2                                          \\ \midrule
                  & Are there any depictions of self-harm in the image?                            & 27.4                                                & 98.9                                          \\ \cline{2-4}
Self-Harm         & Are there visible signs of self-bruising or self-harm through physical trauma? & 74.2                                                & 99.2                                          \\ \cline{2-4}
                  & Does the image contain symbols associated with self-destructive behavior?      & 30.2                                                & 81.4                                          \\ \bottomrule
\end{tabular}%
}
\caption{Quantitative results of VLM on fine-grained moderator questions.}
\label{tab:pass2}
\end{table}

\noindent \textbf{Counterfactual Subobject Explanations for Obfuscation.}
We investigate the ability of CSE to generate a successful counterfactual explanation on an unsafe image \(X\) to satisfy two requirements: (1) the generated counterfactual example \(X'\) must be a convincing representation of another class such that it has a softmax score greater than a threshold $T$ on another class, and (2) the search space that the counterfactual example \(X'\) exists in must be found by searching $N$ or fewer different regions. Since, there are \(2^K\) different combinations of regions to be analyzed in \(X\) with \(K\) number of regions, we limit the search space to a certain number of regions in our evaluation. 
%
In our experiments on unsafe images, we select the threshold for softmax score $T$ to be 0.5 and the threshold for regions to be 10.

\subsection{Results and Discussion}

\noindent \textbf{ConditionalVLM.} The results for the coarse-grained evaluations of the VLM are shown in Table \ref{tab:pass1}. In this table, we present the accuracy of four models, including our model, ConditionalBLIP, that convert images to text, specifically focusing on their ability to identify unsafe attributes in images based on generic questions. In this experiment, a total of 2000 unsafe image samples from each category of unsafe image datasets were tested. The results show that ConditionalBLIP is able to significantly outperform other state-of-the-art models in identifying the unsafe image attributes of unsafe images, simply from asking generic questions on the image, with an average correct identification accuracy of 98\% of unsafe image attributes across the three datasets. Compared to the 50\% accuracy by InstructBLIP, 47\% by mPLUG, and 13.5\% by OFA-Large, we observe that existing models are insufficient for describing unsafe images.\looseness=-1

We present the questions and quantitative results of the fine-grained evaluation of ConditionalBLIP in Table \ref{tab:pass2}. We compare ConditionalBLIP against InstructBLIP, which showed the best coarse-grained results compared to other methods that were evaluated in Table \ref{tab:pass1}. Furthermore, the InstructBLIP model is the most similar in implementation to the ConditionalBLIP model, where the primary difference is the usage of the CIIT in ConditionalBLIP. In Table \ref{tab:pass2}, we present the question posed to the VLM, alongside the detection accuracy of InstructBLIP and ConditionalBLIP on these questions. The fine-grained evaluation shows that image conditioning significantly enhances VLMs ability to understand unsafe images, with an average improvement in accuracy of 38.2\% across the questions. The comparison between the performances of InstructBLIP and ConditionalBLIP reveals significant differences in their respective abilities to identify and describe unsafe content in visual data. For example, given a question on a randomly selected cyberbullying image, specifically asking ``What are the people doing?'', the responses from the two models were noticeably distinct. InstructBLIP stated, ``The people in the image are posing for a photograph,'' a general and incorrect analysis that fails to capture the offensive nature of the image. In contrast, ConditionalBLIP accurately identified the behavior, stating ``In the image, a woman is making an offensive gesture, such as flipping someone off, with her middle finger.'' This response was consistent with the actual content of the image.
%
%
By employing contrastive language-image pre-training and conditioning the language prompt using pre-trained unsafe image classifiers, ConditionalBLIP is able to parse the unsafe visual embedding effectively. 

\noindent \textbf{Counterfactual Subobject Explanations for Obfuscation.}


For the counterfactual image obfuscation experiments, we test on 585 sexually explicit, cyberbullying and self-harm images. We compare our method against the CSRA method, setting numROI = 10 to match time complexity. Previous work showed gradient-based attribution maps were unsuitable for obfuscating unsafe images \cite{bethany2023towards}. Our trained models show improvements of 13.9\% on sexually explicit, 22.0\% on cyberbullying, and 39.5\% on self-harm images when comparing CSRA vs CSE.

\begin{table}[b]
\centering
\resizebox{0.9\columnwidth}{!}{%
\begin{tabular}{lllll}
\toprule
\textbf{Dataset}           & \textbf{Attribution Map} & \textbf{Counterfactual} & \textbf{Avg Depth} & \textbf{Avg Obfuscation} \\ \midrule
                  & \textbf{FullGrad}        & 90.6                     & 5.8      & 35.0              \\ 
                  & Ablation-CAM    & 90.6                     & 5.8      & 35.2              \\ 
Sexually Explicit & Grad-CAM        & 90.6                     & 5.8      & 35.2              \\ 
                  & Grad-CAM++      & 90.6                     & 5.8      & 35.2              \\ 
                  & XGrad-CAM       & 90.6                     & 5.8      & 35.2              \\ \midrule
                  & \textbf{FullGrad}        & 82.0                     & 5.2      & 35.2              \\ 
                  & Ablation-CAM    & 79.5                     & 5.1      & 34.2              \\ 
Cyberbullying     & Grad-CAM        & 79.5                     & 5.1      & 34.2              \\ 
                  & Grad-CAM++      & 79.5                     & 5.1      & 34.2              \\ 
                  & XGrad-CAM       & 79.5                     & 5.1      & 34.2              \\ \midrule
                  & \textbf{FullGrad}        & 72.8                     & 5.6      & 50.1              \\ 
                  & Ablation-CAM    & 72.8                     & 5.6      & 50.1              \\ 
Self-Harm         & Grad-CAM        & 72.8                     & 5.6      & 50.1              \\ 
                  & Grad-CAM++      & 72.8                     & 5.6      & 50.1              \\ 
                  & XGrad-CAM       & 72.8                     & 5.6      & 50.1              \\ \bottomrule
\end{tabular}%
}
\caption{Quantitative results of CSE using different attribution map methods.}
\label{tab:attribution}
\end{table}

\begin{table}[t]
\centering
\resizebox{0.9\columnwidth}{!}{%
\begin{tabular}{lllll}
\toprule
\textbf{Dataset}           & \textbf{Attribution Map} & \textbf{Counterfactual} & \textbf{Avg Depth} & \textbf{Avg Obfuscation} \\ \midrule
                  & \textbf{BASS}         & 90.6                     & 5.8      & 35.0              \\ 
                  & SLIC         & 76.6                     & 7.6      & 33.0              \\ 
Sexually Explicit & Felzenswalb  & 19.9                     & 7.5      & 12.2              \\ 
                  & Watershed    & 51.2                     & 7.9      & 31.9              \\ 
                  & SAM          & 29.5                     & 7.4      & 33.2              \\ \midrule
                  & \textbf{BASS}         & 82.0                     & 5.2      & 35.2              \\ 
                  & SLIC         & 60.0                     & 6.3      & 25.9              \\ 
Cyberbullying     & Felzenswalb  & 20.5                     & 6.3      & 17.6              \\ 
                  & Watershed    & 50.0                     & 6.6      & 23.9              \\ 
                  & SAM          & 50.0                     & 6.6      & 40.2              \\ \midrule
                  & \textbf{BASS}         & 72.8                     & 5.6      & 50.1              \\ 
                  & SLIC         & 33.4                     & 6.6      & 26.3              \\ 
Self-Harm         & Felzenswalb  & 38.4                     & 6.5      & 47.5              \\ 
                  & Watershed    & 33.1                     & 6.8      & 24.6              \\ 
                  & SAM          & 39.5                     & 6.2      & 70.6              \\ \bottomrule
\end{tabular}%
}
\caption{Quantitative results of CSE on different segmentation methods.}
\label{tab:segmentation}
\end{table}

We tested various attribution map methods with BASS \cite{uziel2019bayesian} as the constant segmentation method on unsafe image samples, with results in Table \ref{tab:attribution}. The average search space required to find a counterfactual example was presented, showing that different attribution map methods do not significantly impact CSE, with most generating similar highest average attribution scores in similar areas. The exception was the FullGrad method, which provided slightly more successful counterfactual examples, better average search space, and fewer obfuscated regions. This can be  attributed to FullGrad's more dispersed attributions across the image, which does not restrict the search space as much, and its unique method of satisfying local and global importance by aggregating information from both input-gradient and intermediate bias-gradients, thus aiding CSE in finding suitable counterfactual explanations more readily.\looseness=-1


We tested different segmentation methods with FullGrad as the constant attribution map method on unsafe image samples, and the results are in Table \ref{tab:segmentation}. The choice of segmentation method significantly impacted the number of successful counterfactual explanations, average search space, and average number of regions obfuscated. BASS was the most effective, with a combination of BASS and FullGrad yielding 81.8\% successful counterfactual examples, a search depth of 5.5, and an average of 40.1\% of the image obfuscated. The segmentation's effect on counterfactual examples can be seen in Figure \ref{fig:partitions}, and as Table \ref{tab:segmentation} showed, methods like BASS are key for successful counterfactual explanations, as they break the image into non-intersecting, color and spatially coherent subobjects.

\noindent \textbf{Ablation Study.} 
To evaluate our vision-language model’s conditioning, we conducted an ablation study by changing unsafe classifier guidance on the Image Instruction-guided Transformer or CIIT model’s instruct prompt embedding from 1 to 0.  This conditioning on zero-shot instruct embeddings yielded acceptable results for unsafe images by allowing CIIT to match and parse the unsafe visual embedding effectively, while also providing more explicit control over unsafe visual feature correlation with conditioned instruct prompt. For instance, the LLM decoder's output for an unsafe image changed to suggest ``women are performing potential erotic dance in a bar'' vs. ``women dancing in a bar''. These results suggest that conditioning is a promising approach for vision language models. Further details are in Appendix [ConditioningVLM]. Future research could explore alternative safeguard mechanisms for vision language models.\looseness=-1

\section{Conclusion}
\label{conclusion}

In this work, we have presented ConditionalVLM, a visual reasoning framework that generates accurate rationales for unsafe image descriptions by leveraging state-of-the-art VLMs conditioned on pre-trained unsafe image classifiers, and CSE, a counterfactual visual explanation technique to obfuscate the unsafe regions in unsafe images for safer sharing. We evaluated these two methods on three categories of unsafe images. An implementation of ConditionalVLM, which we called ConditionalBLIP showed superior performance compared to other state-of-the-art image-to-text models on describing unsafe images. We also compare CSE against another recent unsafe image obfuscation method and show how our approach is effective in generating causal explanations for obfuscating unsafe images.

\section*{Acknowledgments}
This research project and the preparation of this publication were funded in part by the Department of Homeland Security (DHS), United States Secret Service, National Computer Forensics Institute (NCFI) via contract number 70US0920D70090004 and by NSF Grant No. 2245983.

\bibliography{aaai24}

\begin{thebibliography}{54}
\providecommand{\natexlab}[1]{#1}

\bibitem[{bbc(2021)}]{bbc2021facebook}
 2021.
\newblock {Facebook moderator: ‘Every day was a nightmare’}.
\newblock \url{https://www.bbc.com/news/technology-57088382}.

\bibitem[{reu(2021)}]{reuters2021judge}
 2021.
\newblock {Judge OKs \$85 mln settlement of Facebook moderators' PTSD claims}.
\newblock \url{https://www.reuters.com/legal/transactional/judge-oks-85-mln-settlement-facebook-moderators-ptsd-claims-2021-07-23/}.

\bibitem[{Vic(2023)}]{Vicuna2023}
 2023.
\newblock {Vicuna}.
\newblock \url{https://github.com/lm-sys/FastChat}.

\bibitem[{Achanta et~al.(2010)Achanta, Shaji, Smith, Lucchi, Fua, and S{\"u}sstrunk}]{achanta2010slic}
Achanta, R.; Shaji, A.; Smith, K.; Lucchi, A.; Fua, P.; and S{\"u}sstrunk, S. 2010.
\newblock Slic superpixels.
\newblock Technical report.

\bibitem[{Adler and Chenoa~Cooper(2022)}]{adler2022tornado}
Adler, R.~A.; and Chenoa~Cooper, S. 2022.
\newblock ``When a Tornado Hits Your Life:'' Exploring Cyber Sexual Abuse Survivors’ Perspectives on Recovery.
\newblock \emph{Journal of Counseling Sexology \& Sexual Wellness: Research, Practice, and Education}, 4(1): 1--8.

\bibitem[{Alayrac et~al.(2022)Alayrac, Donahue, Luc, Miech, Barr, Hasson, Lenc, Mensch, Millican, Reynolds et~al.}]{alayrac2022flamingo}
Alayrac, J.-B.; Donahue, J.; Luc, P.; Miech, A.; Barr, I.; Hasson, Y.; Lenc, K.; Mensch, A.; Millican, K.; Reynolds, M.; et~al. 2022.
\newblock Flamingo: a visual language model for few-shot learning.
\newblock \emph{Advances in Neural Information Processing Systems}, 35: 23716--23736.

\bibitem[{Are(2020)}]{are2020instagram}
Are, C. 2020.
\newblock How Instagram’s algorithm is censoring women and vulnerable users but helping online abusers.
\newblock \emph{Feminist media studies}, 20(5): 741--744.

\bibitem[{Ashurst and McAlinden(2015)}]{ashurst2015young}
Ashurst, L.; and McAlinden, A.-M. 2015.
\newblock Young people, peer-to-peer grooming and sexual offending: Understanding and responding to harmful sexual behaviour within a social media society.
\newblock \emph{Probation Journal}, 62(4): 374--388.

\bibitem[{Berrios et~al.(2023)Berrios, Mittal, Thrush, Kiela, and Singh}]{berrios2023towards}
Berrios, W.; Mittal, G.; Thrush, T.; Kiela, D.; and Singh, A. 2023.
\newblock Towards Language Models That Can See: Computer Vision Through the LENS of Natural Language.
\newblock \emph{arXiv preprint arXiv:2306.16410}.

\bibitem[{Bethany et~al.(2023)Bethany, Seong, Silva, Beebe, Vishwamitra, and Najafirad}]{bethany2023towards}
Bethany, M.; Seong, A.; Silva, S.~H.; Beebe, N.; Vishwamitra, N.; and Najafirad, P. 2023.
\newblock Towards targeted obfuscation of adversarial unsafe images using reconstruction and counterfactual super region attribution explainability.
\newblock In \emph{32nd USENIX Security Symposium (USENIX Security 23)}, 643--660.

\bibitem[{{Billy Perrigo}(2019)}]{facebookalwayshuman}
{Billy Perrigo}. 2019.
\newblock {Facebook Says It’s Removing More Hate Speech Than Ever Before. But There’s a Catch}.

\bibitem[{Binder(2019)}]{facebook_tool_revenge_porn}
Binder, M. 2019.
\newblock Facebook claims its new AI technology can automatically detect revenge porn.
\newblock \url{https://mashable.com/article/facebook-ai-tool-revenge-porn}.

\bibitem[{Bronstein(2021)}]{bronstein2021deplatforming}
Bronstein, C. 2021.
\newblock Deplatforming sexual speech in the age of FOSTA/SESTA.
\newblock \emph{Porn Studies}, 8(4): 367--380.

\bibitem[{Cabral et~al.(2021)Cabral, Haucap, Parker, Petropoulos, Valletti, and Van~Alstyne}]{cabral2021eu}
Cabral, L.; Haucap, J.; Parker, G.; Petropoulos, G.; Valletti, T.~M.; and Van~Alstyne, M.~W. 2021.
\newblock The EU digital markets act: a report from a panel of economic experts.
\newblock \emph{Cabral, L., Haucap, J., Parker, G., Petropoulos, G., Valletti, T., and Van Alstyne, M., The EU Digital Markets Act, Publications Office of the European Union, Luxembourg}.

\bibitem[{Chandrasekaran et~al.(2021)Chandrasekaran, Lei, Kacker, and Kuhn}]{chandrasekaran2021combinatorial}
Chandrasekaran, J.; Lei, Y.; Kacker, R.; and Kuhn, D.~R. 2021.
\newblock A combinatorial approach to explaining image classifiers.
\newblock In \emph{2021 IEEE International Conference on Software Testing, Verification and Validation Workshops (ICSTW)}, 35--43. IEEE.

\bibitem[{Chattopadhay et~al.(2018)Chattopadhay, Sarkar, Howlader, and Balasubramanian}]{chattopadhay2018grad}
Chattopadhay, A.; Sarkar, A.; Howlader, P.; and Balasubramanian, V.~N. 2018.
\newblock Grad-cam++: Generalized gradient-based visual explanations for deep convolutional networks.
\newblock In \emph{2018 IEEE winter conference on applications of computer vision (WACV)}, 839--847. IEEE.

\bibitem[{Chelmis and Yao(2019)}]{chelmis2019minority}
Chelmis, C.; and Yao, M. 2019.
\newblock Minority report: Cyberbullying prediction on Instagram.
\newblock In \emph{Proceedings of the 10th ACM conference on web science}, 37--45.

\bibitem[{Dai et~al.(2023)Dai, Li, Li, Tiong, Zhao, Wang, Li, Fung, and Hoi}]{dai2023instructblip}
Dai, W.; Li, J.; Li, D.; Tiong, A. M.~H.; Zhao, J.; Wang, W.; Li, B.; Fung, P.; and Hoi, S. 2023.
\newblock InstructBLIP: Towards General-purpose Vision-Language Models with Instruction Tuning.
\newblock arXiv:2305.06500.

\bibitem[{Deng et~al.(2009)Deng, Dong, Socher, Li, Li, and Fei-Fei}]{deng2009imagenet}
Deng, J.; Dong, W.; Socher, R.; Li, L.-J.; Li, K.; and Fei-Fei, L. 2009.
\newblock Imagenet: A large-scale hierarchical image database.
\newblock In \emph{2009 IEEE conference on computer vision and pattern recognition}, 248--255. Ieee.

\bibitem[{Exon(1996)}]{exon1996communications}
Exon, J. 1996.
\newblock The Communications Decency Act.
\newblock \emph{Federal Communications Law Journal}, 49(1): 4.

\bibitem[{Fang et~al.(2023)Fang, Wang, Xie, Sun, Wu, Wang, Huang, Wang, and Cao}]{fang2023eva}
Fang, Y.; Wang, W.; Xie, B.; Sun, Q.; Wu, L.; Wang, X.; Huang, T.; Wang, X.; and Cao, Y. 2023.
\newblock Eva: Exploring the limits of masked visual representation learning at scale.
\newblock In \emph{Proceedings of the IEEE/CVF Conference on Computer Vision and Pattern Recognition}, 19358--19369.

\bibitem[{Felzenszwalb and Huttenlocher(2004)}]{felzenszwalb2004efficient}
Felzenszwalb, P.~F.; and Huttenlocher, D.~P. 2004.
\newblock Efficient graph-based image segmentation.
\newblock \emph{International journal of computer vision}, 59(2): 167--181.

\bibitem[{Fu et~al.(2020)Fu, Hu, Dong, Guo, Gao, and Li}]{fu2020axiom}
Fu, R.; Hu, Q.; Dong, X.; Guo, Y.; Gao, Y.; and Li, B. 2020.
\newblock Axiom-based Grad-CAM: Towards Accurate Visualization and Explanation of CNNs.
\newblock In \emph{BMVC}.

\bibitem[{Gesley(2021)}]{gesley2021germany}
Gesley, J. 2021.
\newblock Germany: Network Enforcement Act Amended to Better Fight Online Hate Speech.
\newblock In \emph{Library of Congress, at: https://www. loc. gov/item/global-legal-monito r/2021-07-06/germany-network-enforcement-act-amended-to-better-fight-online-hat e-speech/\#:\~{}: text= Article\% 20Germany\% 3A\% 20Network\% 20Enforcement\% 20Act, fake\% 20news\% 20in\% 20social\% 20networks}.

\bibitem[{Gildenblat and contributors(2021)}]{jacobgilpytorchcam}
Gildenblat, J.; and contributors. 2021.
\newblock PyTorch library for CAM methods.
\newblock \url{https://github.com/jacobgil/pytorch-grad-cam}.

\bibitem[{Hargrave and Livingstone(2009)}]{hargrave2009harm}
Hargrave, A.~M.; and Livingstone, S.~M. 2009.
\newblock Harm and offence in media content: A review of the evidence.

\bibitem[{Hendricks(2021)}]{cyberbullying_increase_12news}
Hendricks, T. 2021.
\newblock Cyberbullying increased 70\% during the pandemic; Arizona schools are taking action.
\newblock \url{https://www.12news.com/article/news/crime/cyberbullying-increased-70-during-the-pandemic-arizona-schools-are-taking-action/75-fadf8d2c-cf11-43f0-b074-5de485a3247d}.

\bibitem[{John et~al.(2018)John, Glendenning, Marchant, Montgomery, Stewart, Wood, Lloyd, Hawton et~al.}]{john2018self}
John, A.; Glendenning, A.~C.; Marchant, A.; Montgomery, P.; Stewart, A.; Wood, S.; Lloyd, K.; Hawton, K.; et~al. 2018.
\newblock Self-harm, suicidal behaviours, and cyberbullying in children and young people: Systematic review.
\newblock \emph{Journal of Medical Internet Research}, 20(4).

\bibitem[{Kim(2021)}]{alagiri2021nsfw}
Kim, A. 2021.
\newblock NSFW Data Scraper.
\newblock \url{https://github.com/alex000kim/nsfw_data_scraper}.

\bibitem[{Kirillov et~al.(2023)Kirillov, Mintun, Ravi, Mao, Rolland, Gustafson, Xiao, Whitehead, Berg, Lo et~al.}]{kirillov2023segment}
Kirillov, A.; Mintun, E.; Ravi, N.; Mao, H.; Rolland, C.; Gustafson, L.; Xiao, T.; Whitehead, S.; Berg, A.~C.; Lo, W.-Y.; et~al. 2023.
\newblock Segment anything.
\newblock \emph{arXiv preprint arXiv:2304.02643}.

\bibitem[{Krause(2009)}]{krause2009identifying}
Krause, M. 2009.
\newblock Identifying and managing stress in child pornography and child exploitation investigators.
\newblock \emph{Journal of Police and Criminal Psychology}, 24(1): 22--29.

\bibitem[{Lenhart, Ybarra, and Price-Feeney(2016)}]{lenhart2016nonconsensual}
Lenhart, A.; Ybarra, M.; and Price-Feeney, M. 2016.
\newblock Nonconsensual image sharing: one in 25 Americans has been a victim of" revenge porn".

\bibitem[{Li et~al.(2022)Li, Xu, Tian, Wang, Yan, Bi, Ye, Chen, Xu, Cao et~al.}]{li2022mplug}
Li, C.; Xu, H.; Tian, J.; Wang, W.; Yan, M.; Bi, B.; Ye, J.; Chen, H.; Xu, G.; Cao, Z.; et~al. 2022.
\newblock mPLUG: Effective and Efficient Vision-Language Learning by Cross-modal Skip-connections.
\newblock In \emph{Proceedings of the 2022 Conference on Empirical Methods in Natural Language Processing}, 7241--7259.

\bibitem[{Li et~al.(2023)Li, Li, Savarese, and Hoi}]{li2023blip}
Li, J.; Li, D.; Savarese, S.; and Hoi, S. 2023.
\newblock Blip-2: Bootstrapping language-image pre-training with frozen image encoders and large language models.
\newblock \emph{arXiv preprint arXiv:2301.12597}.

\bibitem[{Li et~al.(2017)Li, Vishwamitra, Knijnenburg, Hu, and Caine}]{li2017effectiveness}
Li, Y.; Vishwamitra, N.; Knijnenburg, B.~P.; Hu, H.; and Caine, K. 2017.
\newblock Effectiveness and users' experience of obfuscation as a privacy-enhancing technology for sharing photos.
\newblock \emph{Proceedings of the ACM on Human-Computer Interaction}, 1(CSCW): 1--24.

\bibitem[{{Meta}(2022)}]{appealedcontent}
{Meta}. 2022.
\newblock {Appealed Content}.
\newblock https://transparency.fb.com/policies/improving/appealed-content-metric/.

\bibitem[{Neubert and Protzel(2014)}]{neubert2014compact}
Neubert, P.; and Protzel, P. 2014.
\newblock Compact watershed and preemptive slic: On improving trade-offs of superpixel segmentation algorithms.
\newblock In \emph{2014 22nd international conference on pattern recognition}, 996--1001. IEEE.

\bibitem[{Paszke et~al.(2019)Paszke, Gross, Massa, Lerer, Bradbury, Chanan, Killeen, Lin, Gimelshein, Antiga, Desmaison, Kopf, Yang, DeVito, Raison, Tejani, Chilamkurthy, Steiner, Fang, Bai, and Chintala}]{NEURIPS2019_9015}
Paszke, A.; Gross, S.; Massa, F.; Lerer, A.; Bradbury, J.; Chanan, G.; Killeen, T.; Lin, Z.; Gimelshein, N.; Antiga, L.; Desmaison, A.; Kopf, A.; Yang, E.; DeVito, Z.; Raison, M.; Tejani, A.; Chilamkurthy, S.; Steiner, B.; Fang, L.; Bai, J.; and Chintala, S. 2019.
\newblock PyTorch: An Imperative Style, High-Performance Deep Learning Library.
\newblock In Wallach, H.; Larochelle, H.; Beygelzimer, A.; d\textquotesingle Alch\'{e}-Buc, F.; Fox, E.; and Garnett, R., eds., \emph{Advances in Neural Information Processing Systems 32}, 8024--8035. Curran Associates, Inc.

\bibitem[{Radford et~al.(2021)Radford, Kim, Hallacy, Ramesh, Goh, Agarwal, Sastry, Askell, Mishkin, Clark et~al.}]{radford2021learning}
Radford, A.; Kim, J.~W.; Hallacy, C.; Ramesh, A.; Goh, G.; Agarwal, S.; Sastry, G.; Askell, A.; Mishkin, P.; Clark, J.; et~al. 2021.
\newblock Learning transferable visual models from natural language supervision.
\newblock In \emph{International conference on machine learning}, 8748--8763. PMLR.

\bibitem[{Ramaswamy et~al.(2020)}]{ramaswamy2020ablation}
Ramaswamy, H.~G.; et~al. 2020.
\newblock Ablation-cam: Visual explanations for deep convolutional network via gradient-free localization.
\newblock In \emph{Proceedings of the IEEE/CVF Winter Conference on Applications of Computer Vision}, 983--991.

\bibitem[{Ramesh et~al.(2022)Ramesh, Dhariwal, Nichol, Chu, and Chen}]{ramesh2022hierarchical}
Ramesh, A.; Dhariwal, P.; Nichol, A.; Chu, C.; and Chen, M. 2022.
\newblock Hierarchical text-conditional image generation with clip latents.
\newblock \emph{arXiv preprint arXiv:2204.06125}.

\bibitem[{Sanchez et~al.(2019)Sanchez, Grajeda, Baggili, and Hall}]{sanchez2019practitioner}
Sanchez, L.; Grajeda, C.; Baggili, I.; and Hall, C. 2019.
\newblock A practitioner survey exploring the value of forensic tools, AI, filtering, \& safer presentation for investigating child sexual abuse material (CSAM).
\newblock \emph{Digital Investigation}, 29.

\bibitem[{Selvaraju et~al.(2017)Selvaraju, Cogswell, Das, Vedantam, Parikh, and Batra}]{selvaraju2017grad}
Selvaraju, R.~R.; Cogswell, M.; Das, A.; Vedantam, R.; Parikh, D.; and Batra, D. 2017.
\newblock Grad-cam: Visual explanations from deep networks via gradient-based localization.
\newblock In \emph{Proceedings of the IEEE international conference on computer vision}, 618--626.

\bibitem[{Srinivas and Fleuret(2019)}]{srinivas2019full}
Srinivas, S.; and Fleuret, F. 2019.
\newblock Full-gradient representation for neural network visualization.
\newblock \emph{Advances in neural information processing systems}, 32.

\bibitem[{Steiger et~al.(2021)Steiger, Bharucha, Venkatagiri, Riedl, and Lease}]{steiger2021psychological}
Steiger, M.; Bharucha, T.~J.; Venkatagiri, S.; Riedl, M.~J.; and Lease, M. 2021.
\newblock The psychological well-being of content moderators: the emotional labor of commercial moderation and avenues for improving support.
\newblock In \emph{Proceedings of the 2021 CHI conference on human factors in computing systems}, 1--14.

\bibitem[{Tenbarge(2023)}]{instagramrationalemoderator}
Tenbarge, K. 2023.
\newblock {Instagram's sex censorship sweeps up educators, adult stars and sex workers}.

\bibitem[{Touvron et~al.(2023)Touvron, Lavril, Izacard, Martinet, Lachaux, Lacroix, Rozi{\`e}re, Goyal, Hambro, Azhar et~al.}]{touvron2023llama}
Touvron, H.; Lavril, T.; Izacard, G.; Martinet, X.; Lachaux, M.-A.; Lacroix, T.; Rozi{\`e}re, B.; Goyal, N.; Hambro, E.; Azhar, F.; et~al. 2023.
\newblock Llama: Open and efficient foundation language models.
\newblock \emph{arXiv preprint arXiv:2302.13971}.

\bibitem[{Uziel, Ronen, and Freifeld(2019)}]{uziel2019bayesian}
Uziel, R.; Ronen, M.; and Freifeld, O. 2019.
\newblock Bayesian adaptive superpixel segmentation.
\newblock In \emph{Proceedings of the IEEE/CVF International Conference on Computer Vision}, 8470--8479.

\bibitem[{van~der Walt et~al.(2014)van~der Walt, {S}ch\"onberger, {Nunez-Iglesias}, {B}oulogne, {W}arner, {Y}ager, {G}ouillart, {Y}u, and the scikit-image contributors}]{scikit-image}
van~der Walt, S.; {S}ch\"onberger, J.~L.; {Nunez-Iglesias}, J.; {B}oulogne, F.; {W}arner, J.~D.; {Y}ager, N.; {G}ouillart, E.; {Y}u, T.; and the scikit-image contributors. 2014.
\newblock scikit-image: image processing in {P}ython.
\newblock \emph{PeerJ}, 2: e453.

\bibitem[{Vermeire et~al.(2022)Vermeire, Brughmans, Goethals, de~Oliveira, and Martens}]{vermeire2022explainable}
Vermeire, T.; Brughmans, D.; Goethals, S.; de~Oliveira, R. M.~B.; and Martens, D. 2022.
\newblock Explainable image classification with evidence counterfactual.
\newblock \emph{Pattern Analysis and Applications}, 25(2): 315--335.

\bibitem[{Vishwamitra et~al.(2021)Vishwamitra, Hu, Luo, and Cheng}]{vishwamitra2021towards}
Vishwamitra, N.; Hu, H.; Luo, F.; and Cheng, L. 2021.
\newblock Towards Understanding and Detecting Cyberbullying in Real-world Images.
\newblock In \emph{2020 19th IEEE International Conference on Machine Learning and Applications (ICMLA)}.

\bibitem[{Wachter, Mittelstadt, and Russell(2017)}]{wachter2017counterfactual}
Wachter, S.; Mittelstadt, B.; and Russell, C. 2017.
\newblock Counterfactual explanations without opening the black box: Automated decisions and the GDPR.
\newblock \emph{Harv. JL \& Tech.}, 31: 841.

\bibitem[{Wang et~al.(2022)Wang, Yang, Men, Lin, Bai, Li, Ma, Zhou, Zhou, and Yang}]{wang2022ofa}
Wang, P.; Yang, A.; Men, R.; Lin, J.; Bai, S.; Li, Z.; Ma, J.; Zhou, C.; Zhou, J.; and Yang, H. 2022.
\newblock Ofa: Unifying architectures, tasks, and modalities through a simple sequence-to-sequence learning framework.
\newblock In \emph{International Conference on Machine Learning}, 23318--23340. PMLR.

\bibitem[{Wang et~al.(2023)Wang, Bao, Dong, Bjorck, Peng, Liu, Aggarwal, Mohammed, Singhal, Som et~al.}]{wang2023image}
Wang, W.; Bao, H.; Dong, L.; Bjorck, J.; Peng, Z.; Liu, Q.; Aggarwal, K.; Mohammed, O.~K.; Singhal, S.; Som, S.; et~al. 2023.
\newblock Image as a Foreign Language: BEiT Pretraining for Vision and Vision-Language Tasks.
\newblock In \emph{Proceedings of the IEEE/CVF Conference on Computer Vision and Pattern Recognition}, 19175--19186.

\end{thebibliography}

\clearpage

\onecolumn

\end{document}